\documentclass[11pt,a4paper]{article}
\usepackage[hyperref]{naaclhlt2018}
\usepackage{times}
\usepackage{latexsym}

\usepackage{url}
\aclfinalcopy

\usepackage{graphicx} 
\usepackage{multicol}
\usepackage{subfigure} 
\usepackage{booktabs} 
\usepackage{algorithm} 
\usepackage{algorithmic} 
\usepackage{color}
\usepackage{amsmath}
\usepackage{amssymb}
\usepackage{amsfonts}
\usepackage{multirow, varwidth}
\usepackage{stfloats}
\usepackage{verbatim}
\makeatletter
\newif\if@restonecol
\makeatother
\usepackage{xr}
\usepackage[amsmath,thmmarks]{ntheorem}

\newcommand{\argmax}{\operatornamewithlimits{argmax}}

\DeclareRobustCommand\onedot{\futurelet\@let@token\@onedot}
\def\onedot{. }
\def\eg{\emph{e.g}\onedot}

\usepackage{color}

\title{Multimodal Named Entity Recognition for Short Social Media Posts}

\author{Seungwhan Moon\textsuperscript{1,2}, Leonardo Neves\textsuperscript{2}, Vitor Carvalho\textsuperscript{3}\\
\textsuperscript{1} Language Technologies Institute, Carnegie Mellon University\\
\textsuperscript{2} Snap Research\\
\textsuperscript{3} Intuit\\
\texttt{seungwhm@cs.cmu.edu, lneves@snap.com, vitor\_carvalho@intuit.com}
}

\date{}

\begin{document}
\maketitle

\begin{abstract}
We introduce a new task called Multimodal Named Entity Recognition (MNER) for noisy user-generated data such as tweets or Snapchat captions, which comprise short text with accompanying images.
These social media posts often come in inconsistent or incomplete syntax and lexical notations with very limited surrounding textual contexts, bringing significant challenges for NER.
To this end, we create a new dataset for MNER called SnapCaptions (Snapchat image-caption pairs submitted to public and crowd-sourced stories with fully annotated named entities).
We then build upon the state-of-the-art Bi-LSTM word/character based NER models with 1) a deep image network which incorporates relevant visual context to augment textual information, and 2) a generic \textit{modality-attention} module which learns to attenuate irrelevant modalities while amplifying the most informative ones to extract contexts from, adaptive to each sample and token.
The proposed MNER model with modality attention significantly outperforms the state-of-the-art text-only NER models by successfully leveraging provided visual contexts, opening up potential applications of MNER on myriads of social media platforms.
\end{abstract}

\section{Introduction}
\label{sec:introduction}

\begin{figure}[t]
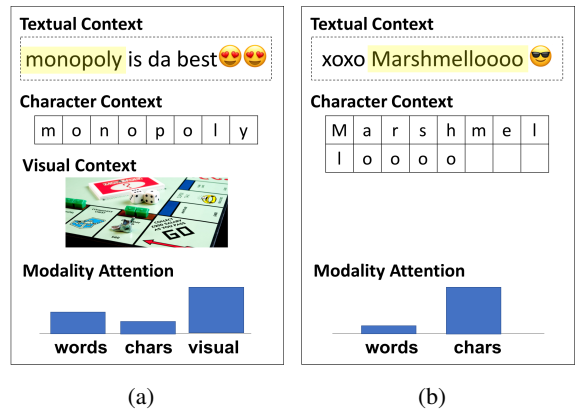

 \include{tex_modules/fig_ner_teasers}
  \vspace{-9pt} 
 \caption{ \textbf{Multimodal NER + modality attention}. (a) Visual contexts help recognizing polysemous entity names (`Monopoly' as in a board game versus an economics term). (b) Modality attention successfully suppresses word embeddings of a unknown token (`Marshmelloooo' with erroneously trailing `o's), and focuses on character-based context (\textit{e.g.} capitalized first letter, and lexical similarity to a known named entity (`Marshmello', a music producer)) for correct prediction. }
  \label{fig:ner_teasers}
  \vspace{-12pt}
\end{figure}

Social media with abundant user-generated posts provide a rich platform for understanding events, opinions and preferences of groups and individuals.
These insights are primarily hidden in unstructured forms of social media posts, such as in free-form text or images without tags.
Named entity recognition (NER), the task of recognizing named entities from free-form text, is thus a critical step for building structural information, allowing for its use in personalized assistance, recommendations, advertisement, etc.

While many previous approaches \cite{Lample+16,Ma+16,Chiu+15,Huang+15} on NER have shown success for well-formed text in recognizing named entities via word context resolution (\textit{e.g.} LSTM with word embeddings) combined with character-level features (\textit{e.g.} CharLSTM/CNN), several additional challenges remain for recognizing named entities from extremely short and coarse text found in social media posts.
For instance, short social media posts often do not provide enough textual contexts to resolve polysemous entities (\textit{e.g.} ``\underline{monopoly} is da best \includegraphics[height=1em]{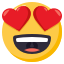}", where `monopoly' may refer to a board game (named entity) or a term in economics).
In addition, noisy text includes a huge number of unknown tokens due to inconsistent lexical notations and frequent mentions of various newly trending entities (\textit{e.g.} ``xoxo \underline{Marshmelloooo} \includegraphics[height=1em]{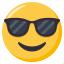}", where `Marshmelloooo' is a mis-spelling of a known entity `Marshmello', a music producer), making word embeddings based neural networks NER models vulnerable.

To address the challenges above for social media posts, we build upon the state-of-the-art neural architecture for NER with the following two novel approaches (Figure \ref{fig:ner_teasers}).
First, we propose to leverage auxiliary modalities for additional context resolution of entities.
For example, many popular social media platforms now provide ways to compose a post in multiple modalities - specifically image and text (\textit{e.g.} \textit{Snapchat} captions, \textit{Twitter} posts with image URLs), from which we can obtain additional context for understanding posts.
While ``\underline{monopoly}" in the previous example is ambiguous in its textual form, an accompanying snap image of a board game can help disambiguate among polysemous entities, thereby correctly recognizing it as a named entity.

Second, we also propose a general modality attention module which chooses per decoding step the most informative modality among available ones (in our case, word embeddings, character embeddings, or visual features) to extract context from.
For example, the modality attention module lets the decoder attenuate the word-level signals for unknown word tokens (\eg ``\underline{Marshmellooooo}" with trailing `o's) and amplifies character-level features intsead (\eg capitalized first letter, lexical similarity to other known named entity token `Marshmello', etc.), thereby suppressing noise information (``UNK" token embedding) in decoding steps.
Note that most of the previous literature in NER or other NLP tasks combine word and character-level information with naive concatenation, which is vulnerable to noisy social media posts.
When an auxiliary image is available, the modality attention module determines to amplify this visual context \eg in disambiguating polysemous entities, or to attenuate visual contexts when they are irrelevant to target named entities, \eg selfies, etc.
Note that the proposed modality attention module is distinct from how attention is used in other sequence-to-sequence literature (\textit{e.g.} attending to a specific token within an input sequence).
Section \ref{sec:related_work} provides the detailed literature review.

\textbf{Our contributions} are three-fold: we propose (1) an LSTM-CNN hybrid multimodal NER network that takes as input both image and text for recognition of a named entity in text input. To the best of our knowledge, our approach is the first work to incorporate visual contexts for named entity recognition tasks.
(2) We propose a general \textit{modality attention} module that selectively chooses modalities to extract primary context from, maximizing information gain and suppressing irrelevant contexts from each modality (we treat words, characters, and images as separate modalities).
(3) We show that the proposed approaches outperform the state-of-the-art NER models (\textbf{both} with and without using additional visual contexts) on our new MNER dataset \textit{SnapCaptions}, a large collection of informal and extremely short social media posts paired with unique images.

\section{Related Work}
\label{sec:related_work}

\textbf{Neural models for NER} have been recently proposed, producing state-of-the-art performance on standard NER tasks.
For example, some of the end-to-end NER systems  \cite{Passos+14,Chiu+15,Huang+15,Lample+16,Ma+16} use a recurrent neural network usually with a CRF \cite{CRF,McCallum+03} for sequence labeling, accompanied with feature extractors for words and characters (CNN, LSTMs, etc.), and achieve the state-of-the-art performance mostly without any use of gazetteers information.
Note that most of these work aggregate textual contexts via concatenation of word embeddings and character embeddings.
Recently, several work have addressed the NER task specifically on noisy short text segments such as Tweets, etc. \cite{Baldwin+15,Aguilar+17}.
They report performance gains from leveraging external sources of information such as lexical information (\eg POS tags, etc.) and/or from several preprocessing steps (\eg token substitution, etc.).
Our model builds upon these state-of-the-art neural models for NER tasks, and improves the model in two critical ways: (1) incorporation of visual contexts to provide auxiliary information for short media posts, and (2) addition of the modality attention module, which better incorporates word embeddings and character embeddings, especially when there are many missing tokens in the given word embedding matrix.
Note that we do not explore the use of gazetteers information or other auxiliary information (POS tags, etc.) \cite{Ratinov+09} as it is not the focus of our study.

\textbf{Attention} modules are widely applied in several deep learning tasks \cite{Xu+15,Chan+15,Sukhbaatar+15,Yao+15}.
For example, they use an attention module to attend to a subset within a single input (a part/region of an image, a specific token in an input sequence of tokens, etc.) at each decoding step in an encoder-decoder framework for image captioning tasks, etc.
~\cite{reiattending2016} explore various attention mechanisms in NLP tasks, but do not incorporate visual components or investigate the impact of such models on noisy social media data.
~\cite{Moon+17a} propose to use attention for a subset of discrete source samples in transfer learning settings.
Our modality attention differs from the previous approaches in that we attenuate or amplifies each modality input as a whole among multiple available modalities, and that we use the attention mechanism essentially to map heterogeneous modalities in a single joint embedding space.
Our approach also allows for re-use of the same model for predicting labels even when some of the modalities are missing in input, as other modalities would still preserve the same semantics in the embeddings space.

\textbf{Multimodal learning} is studied in various domains and applications, aimed at building a joint model that extracts contextual information from multiple modalities (views) of parallel datasets. 

The most relevant task to our multimodal NER system is the task of multimodal machine translation \cite{Elliott+15,Specia+16}, which aims at building a better machine translation system by taking as input a sentence in a source language as well as a corresponding image.
Several standard sequence-to-sequence architectures are explored (\eg a target-language LSTM decoder that takes as input an image first).

Other previous literature include study of Canonical Correlation Analysis (CCA) \cite{Dhillon+11} to learn feature correlations among multiple modalities, which is widely used in many applications.
Other applications include image captioning \cite{Xu+15}, audio-visual recognition \cite{Moon+15a}, visual question answering systems \cite{vqa}, etc.

To the best of our knowledge, our approach is the first work to incorporate visual contexts for named entity recognition tasks.

\section{Proposed Methods}
\label{sec:methods}

\begin{figure}[t]
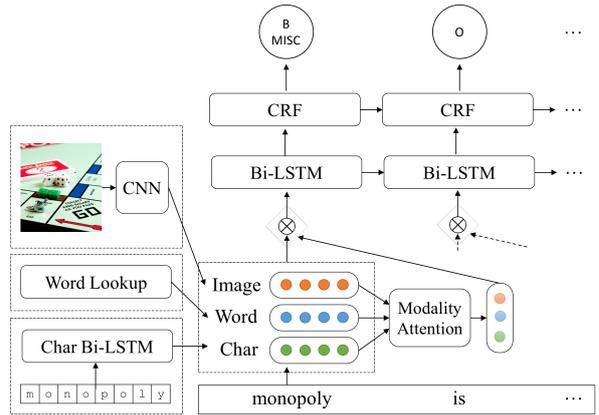

 \include{tex_modules/fig_ner_architecture}
  \vspace{-12pt} 
 \caption{The main architecture for our multimodal NER (MNER) network with modality attention. At each decoding step, word embeddings, character embeddings, and visual features are merged with modality attention. Bi-LSTM/CRF takes as input each token and produces an entity label. }
  \label{fig:ner_architecture}
  \vspace{-6pt}
\end{figure}

Figure \ref{fig:ner_architecture} illustrates the proposed multimodal NER (MNER) model. 
First, we obtain word embeddings, character embeddings, and visual features (Section \ref{subsec:methods:features}).
A Bi-LSTM-CRF model then takes as input a sequence of tokens, each of which comprises a word token, a character sequence, and an image, in their respective representation (Section \ref{subsec:methods:entity_lstm}). 
At each decoding step, representations from each modality are combined via the modality attention module to produce an entity label for each token (\ref{subsec:methods:modality_attention}).
We formulate each component of the model in the following subsections.

\textbf{Notations}:
Let $\mathbf x = \{ \mathbf{x}_t\}_{t=1}^{T}$ a sequence of input tokens with length $T$, with a corresponding label sequence $\mathbf y = \{ \mathbf{y}_t\}_{t=1}^{T}$ indicating named entities (\textit{e.g.} in standard BIO formats).
Each input token is composed of three modalities: $\mathbf{x}_t = \{\mathbf{x}_{t}^{(w)}, \mathbf{x}_{t}^{(c)}, \mathbf{x}_{t}^{(v)}\}$ for word embeddings, character embeddings, and visual embeddings representations, respectively.

\subsection{Features}
\label{subsec:methods:features}

Similar to the state-of-the-art NER approaches \cite{Lample+16,Ma+16,Aguilar+17,Passos+14,Chiu+15,Huang+15}, we use both word embeddings and character embeddings.

\textbf{Word embeddings} are obtained from an unsupervised learning model that learns co-occurrence statistics of words from a large external corpus, yielding word embeddings as distributional semantics \cite{Mikolov+13}.
Specifically, we use pre-trained embeddings from GloVE  \cite{glove}.

\textbf{Character embeddings} are obtained from a Bi-LSTM which takes as input a sequence of characters of each token, similarly to \cite{Lample+16}.
An alternative approach for obtaining character embeddings is using a convolutional neural network as in \cite{Ma+16}, but we find that Bi-LSTM representation of characters yields empirically better results in our experiments.

\textbf{Visual embeddings}: To extract features from an image, we take the final hidden layer representation of a modified version of the convolutional network model called Inception (GoogLeNet) \cite{inceptionv1,inceptionv3} trained on the ImageNet dataset \cite{imagenet} to classify multiple objects in the scene.
Our implementation of the Inception model has deep 22 layers, training of which is made possible via ``network in network" principles and several dimension reduction techniques to improve computing resource utilization.
The final layer representation encodes discriminative information describing what objects are shown in an image, which provide auxiliary contexts for understanding textual tokens and entities in accompanying captions.

Incorporating this visual information onto the traditional NER system is an open challenge, and multiple approaches can be considered.
For instance, one may provide visual contexts only as an initial input to decoder as in some encoder-decoder image captioning systems ~\cite{Vinyals+15}.
However, we empirically observe that an NER decoder which takes as input the visual embeddings at every decoding step (Section \ref{subsec:methods:entity_lstm}), combined with the modality attention module (Section \ref{subsec:methods:modality_attention}), yields better results.

Lastly, we add a transform layer for each feature \eg $\mathbf{x}_{t}^{(w)}, \mathbf{x}_{t}^{(c)}, \mathbf{x}_{t}^{(v)} := \sigma_w(\mathbf{x}_{t}^{(w)}), \sigma_c(\mathbf{x}_{t}^{(c)}), \sigma_v(\mathbf{x}_{t}^{(v)})$ before it is fed to the NER entity LSTM.

\subsection{Bi-LSTM + CRF for Multimodal NER}
\label{subsec:methods:entity_lstm}

Our MNER model is built on a Bi-LSTM and CRF hybrid model.
We use the following implementation for the entity Bi-LSTM.
\vspace{-7pt}
\begin{align}
\label{eq:entity_lstm}
\nonumber \mathbf{i}_t &= \sigma(\mathbf{W}_{xi}\mathbf{h}_{t-1} + \mathbf{W}_{ci}\mathbf{c}_{t-1}) \\
\nonumber \mathbf{c}_t &= (1-\mathbf{i}_t) \odot \mathbf{c}_{t-1} \\
\vspace{-4pt} \nonumber &+ \mathbf{i}_t \odot \text{tanh}(\mathbf{W}_{xc}\overline{\mathbf{x}_t} + \mathbf{W}_{hc}\mathbf{h}_{t-1}) \\
\nonumber \mathbf{o}_t &= \sigma(\mathbf{W}_{xo}\overline{\mathbf{x}_t} + \mathbf{W}_{ho}\mathbf{h}_{t-1} + \mathbf{W}_{co}\mathbf{c}_t) \\
          \mathbf{h}_t &= \text{LSTM}(\overline{\mathbf{x}_t}) \\
\nonumber              &= \mathbf{o}_t \odot \text{tanh}(\mathbf{c}_t)
\end{align}
\noindent where $\overline{\mathbf{x}_t}$ is a weighted average of three modalities $\mathbf{x}_t = \{\mathbf{x}_{t}^{(w)}; \mathbf{x}_{t}^{(c)}; \mathbf{x}_{t}^{(v)}\}$ via the modality attention module, which will be defined in Section \ref{subsec:methods:modality_attention}.
Bias terms for gates are omitted here for simplicity of notation.

We then obtain bi-directional entity token representations $\overleftrightarrow{\mathbf{h}_t} = [\overrightarrow{\mathbf{h}_t}; \overleftarrow{\mathbf{h}_t}]$ by concatenating its left and right context representations. 
To enforce structural correlations between labels in sequence decoding, $\overleftrightarrow{\mathbf{h}_t}$ is then passed to a conditional random field (CRF) to produce a label for each token maximizing the following objective.
\begin{align}
\label{eq:entity_crf}
& \hspace{25pt} \mathbf{y^*} = \argmax_{\mathbf{y}} p(\mathbf{y}|\overleftrightarrow{\mathbf{h}}; \mathbf{W}_{\text{CRF}}) \\
\nonumber &p(\mathbf{y}|\overleftrightarrow{\mathbf{h}}; \mathbf{W}_{\text{CRF}}) = \frac{\prod_{t} \psi_t (\mathbf{y}_{t-1},\mathbf{y}_{t};\overleftrightarrow{\mathbf{h}}) }{ \sum_{\mathbf{y'}} \prod_{t} \psi_t (\mathbf{y'}_{t-1},\mathbf{y'}_{t};\overleftrightarrow{\mathbf{h}}) }
\end{align}
\noindent where $\psi_t(\mathbf{y'},\mathbf{y'};\overleftrightarrow{\mathbf{h}})$ is a potential function, $\mathbf{W}_{\text{CRF}}$ is a set of parameters that defines the potential functions and weight vectors for label pairs ($\mathbf{y'},\mathbf{y'}$).
Bias terms are omitted for brevity of formulation.

The model can be trained via log-likelihood maximization for the training set $\{(\mathbf{x}_i,\mathbf{y}_i)\}$:
\vspace{-10pt}
\begin{align}
\label{eq:crf_loss}
    \mathcal L(\mathbf{W}_{\text{CRF}}) = \sum_i \log p(\mathbf{y}|\overleftrightarrow{\mathbf{h}}; \mathbf{W})
\end{align}
\vspace{-17pt}
\subsection{Modality Attention}
\label{subsec:methods:modality_attention}

The modality attention module learns a unified representation space for multiple available modalities (\eg words, characters, images, etc.), and produces a single vector representation with aggregated knowledge among multiple modalities, based on their weighted importance.
We motivate this module from the following observations.

A majority of the previous literature combine the word and character-level contexts by simply concatenating the word and character embeddings at each decoding step, \textit{e.g.} $\mathbf{h}_t = \text{LSTM}([\mathbf{x}_{t}^{(w)}; \mathbf{x}_{t}^{(c)}])$ in Eq.\ref{eq:entity_lstm}.
However, this naive concatenation of two modalities (word and characters) results in inaccurate decoding, specifically for unknown word token embeddings (\eg an all-zero vector $\mathbf{x}_{t}^{(w)}=\mathbf{0}$ or a random vector $\mathbf{x}_{t}^{(w)}=\boldsymbol{\epsilon} \sim U(-\sigma,+\sigma)$ is assigned for any unknown token $\mathbf{x}_{t}$, thus $\mathbf{h}_t = \text{LSTM}([\mathbf{0}; \mathbf{x}_{t}^{(c)}])$ or $\text{LSTM}([\epsilon; \mathbf{x}_{t}^{(c)}])$).
While this concatenation approach does not cause significant errors for well-formatted text, we observe that it induces performance degradation for our social media post datasets which contain a significant number of missing tokens.

Similarly, naive merging of textual and visual information (\eg $\mathbf{h}_t = \text{LSTM}([\mathbf{x}_{t}^{(w)}; \mathbf{x}_{t}^{(c)}; \mathbf{x}_{t}^{(v)}])$) yields suboptimal results as each modality is treated equally informative, whereas in our datasets some of the images may contain irrelevant contexts to textual modalities.
Hence, ideally there needs a mechanism in which the model can effectively turn the \textit{switch} on and off the modalities adaptive to each sample.

To this end, we propose a general modality attention module, which adaptively attenuates or emphasizes each modality as a whole at each decoding step $t$, and produces a soft-attended context vector $\overline{\mathbf{x}_t}$ as an input token for the entity LSTM.
\begin{align}
\label{eq:modality_attention}
\nonumber [\mathbf{a}_{t}^{(w)},\mathbf{a}_{t}^{(c)},&\mathbf{a}_{t}^{(v)}] = \sigma \big(\mathbf{W}_{m}\cdot[\mathbf{x}_{t}^{(w)}; \mathbf{x}_{t}^{(c)}; \mathbf{x}_{t}^{(v)}] + \mathbf{b}_m \big) \\
\nonumber \alpha_{t}^{(m)} &= \frac{\exp(\mathbf{a}_{t}^{(m)})}{\sum\limits _{m'\in\{w,c,v\}}\hspace{-12pt}\exp(\mathbf{a}_{t}^{(m')})} \hspace{8pt} \forall m \in \{w,c,v\}\\
\overline{\mathbf{x}_t} &= \sum\limits_{m\in\{w,c,v\}} \alpha_{t}^{(m)}\mathbf{x}_{t}^{(m)}
\end{align}

\noindent where $\mathbf{\alpha}_t = [\alpha_{t}^{(w)};\alpha_{t}^{(c)};\alpha_{t}^{(v)}] \in \mathbb{R}^3$ is an attention vector at each decoding step $t$, and $\overline{\mathbf{x}_t}$ is a final context vector at $t$ that maximizes information gain for $\mathbf{x}_t$.
Note that the optimization of the objective function (Eq.\ref{eq:entity_lstm}) with modality attention (Eq.\ref{eq:modality_attention}) requires each modality to have the same dimension (\eg $\mathbf{x}_{t}^{(w)}, \mathbf{x}_{t}^{(c)}, \mathbf{x}_{t}^{(v)} \in \mathbb{R}^p $), and that the transformation via $\mathbf{W}_{m}$ essentially enforces each modality to be mapped into the same unified subspace, where the weighted average of which encodes discrimitive features for recognition of named entities.

When visual context is not provided with each token (as in the traditional NER task), we can define the modality attention for word and character embeddings only in a similar way:
\begin{align}
\label{eq:modality_attention_without_vis}
[\mathbf{a}_{t}^{(w)},\mathbf{a}_{t}^{(c)}] &= \sigma \big(\mathbf{W}_{m}\cdot[\mathbf{x}_{t}^{(w)}; \mathbf{x}_{t}^{(c)}] + \mathbf{b}_m \big) \\
\nonumber \alpha_{t}^{(m)} &= \frac{\exp(\mathbf{a}_{t}^{(m)})}{\sum\limits _{m'\in\{w,c\}}\hspace{-12pt}\exp(\mathbf{a}_{t}^{(m')})} \hspace{8pt} \forall m \in \{w,c\}\\
\nonumber \overline{\mathbf{x}_t} &= \sum\limits_{m\in\{w,c\}} \alpha_{t}^{(m)}\mathbf{x}_{t}^{(m)}
\end{align}

Note that while we apply this modality attention module to the Bi-LSTM+CRF architecture (Section \ref{subsec:methods:entity_lstm}) for its empirical superiority, the module itself is flexible and thus can work with other NER architectures or for other multimodal applications.

\section{Empirical Evaluation}
\label{sec:empirical_evaluation}

\begin{table*}[t]
    \begin{center}
    \scalebox{0.83}{
    \setlength\tabcolsep{3.5pt}
    \begin{tabular}{llcccccc}
    \toprule[\heavyrulewidth]
    \multicolumn{1}{c}{\multirow{2}{*}{Modalities}} & \multicolumn{1}{c}{\multirow{2}{*}{ Model }} & \multicolumn{3}{c}{4 Entity Types (\%)} & \multicolumn{3}{c}{Segmentation (\%)} \\
    \cmidrule(r){3-5}
    \cmidrule(r){6-8}
     &  & Prec. & Recall & F1 & Prec. & Recall & F1  \\
    \midrule
    \midrule
    C & Bi-LSTM/CRF + Bi-CharLSTM  & 5.0 & 28.1 & 8.5 & 68.6 & 10.8 & 18.6 \\
    W & Bi-LSTM/CRF & 38.2 & 53.3 & 44.6 & 82.5 & 50.1 & 62.4 \\    
    \midrule
    W + C & \cite{Aguilar+17} & 45.9 & 48.9 & 47.4 & 74.0 & 61.7 & 67.3 \\
    W + C & \cite{Ma+16} & 46.0 & 51.9 & 48.7 & 76.8 & 61.0 & 68.0 \\
    W + C & \cite{Lample+16} & 47.7 & 49.9 & 48.8 & 74.4 & 63.3 & 68.4 \\   
    W + C & Bi-LSTM/CRF + Bi-CharLSTM w/ Modality Attention & 49.4 & 51.7 & 50.5 & 75.7 & 63.3 & 68.9 \\
    \midrule
    W + C + V & Bi-LSTM/CRF + Bi-CharLSTM + Inception & \textbf{50.5} & 52.3 & 51.4 & 71.9 & \textbf{66.5} & \textbf{69.1} \\    
    W + C + V & Bi-LSTM/CRF + Bi-CharLSTM + Inception w/ Modality Attention & 48.7 & \textbf{58.7} & \textbf{52.4} & \textbf{77.4} & 60.6 & 68.0 \\    
    \bottomrule[\heavyrulewidth]
    \end{tabular}
    }
\end{center}
    \vspace{-7pt}
   \caption{NER performance on the \textit{SnapCaptions} dataset with varying modalities (W: word, C: char, V: visual). We report precision, recall, and F1 score for both entity types recognition (PER, LOC, ORG, MISC) and entity segmentation (untyped recognition - named entity or not) tasks.}
   \vspace{-6pt}
    \label{tab:ner_snap_captions}
\end{table*}

\subsection{SnapCaptions Dataset}
\label{subsec:empirical_evaluation:datasets}

The \textbf{SnapCaptions} dataset is composed of 10K user-generated image (snap) and textual caption pairs where named entities in captions are manually labeled by expert human annotators (entity types: PER, LOC, ORG, MISC).
These captions are collected exclusively from snaps submitted to public and crowd-sourced stories (aka Snapchat \textit{Live Stories} or \textit{Our Stories}).
Examples of such public crowd-sourced stories are ``New York Story'' or ``Thanksgiving Story'', which comprise snaps that are aggregated for various public events, venues, etc.
All snaps were posted between year 2016 and 2017, and do not contain raw images or other associated information (only textual captions and obfuscated visual descriptor features extracted from the pre-trained InceptionNet are available). 
We split the dataset into train (70\%), validation (15\%), and test sets (15\%).
The captions data have average length of 30.7 characters (5.81 words) with vocabulary size 15,733, where 6,612 are considered unknown tokens from Stanford GloVE embeddings \cite{glove}.
Named entities annotated in the SnapCaptions dataset include many of new and emerging entities, and they are found in various surface forms (various nicknames, typos, etc.)
To the best of our knowledge, \textit{SnapCaptions} is the only dataset that contains natural image-caption pairs with expert-annotated named entities.

\subsection{Baselines}
\label{subsec:empirical_evaluation:baselines}

\textbf{Task}: given a caption and a paired image (if used), the goal is to label every token in a caption in BIO scheme (B: beginning, I: inside, O: outside) \cite{bio}.
We report the performance of the following state-of-the-art NER models as baselines, as well as several configurations of our proposed approach to examine contributions of each component (W: word, C: char, V: visual).

\begin{itemize}
    \item Bi-LSTM/CRF (W only): only takes word token embeddings (Stanford GloVE) as input. The rest of the architecture is kept the same.
    \item Bi-LSTM/CRF + Bi-CharLSTM (C only): only takes a character sequence of each word token as input. (No word embeddings)
    \item Bi-LSTM/CRF + Bi-CharLSTM (W+C) \cite{Lample+16}: takes as input both word embeddings and character embeddings extracted from a Bi-CharLSTM. Entity LSTM takes concatenated vectors of word and character embeddings as input tokens.
    \item Bi-LSTM/CRF + CharCNN (W+C) \cite{Ma+16}: uses character embeddings extracted from a CNN instead.
    \item Bi-LSTM/CRF + CharCNN (W+C) + Multi-task \cite{Aguilar+17}: trains the model to perform both recognition (into multiple entity types) as well as segmentation (binary) tasks.
    \item (\textbf{proposed}) Bi-LSTM/CRF + Bi-CharLSTM with modality attention (W+C): uses the modality attention to merge word and character embeddings.
    \item (\textbf{proposed}) Bi-LSTM/CRF + Bi-CharLSTM + Inception (W+C+V): takes as input visual contexts extracted from InceptionNet as well, concatenated with word and char vectors.
    \item (\textbf{proposed}) Bi-LSTM/CRF + Bi-CharLSTM + Inception with modality attention (W+C+V): uses the modality attention to merge word, character, and visual embeddings as input to entity LSTM.
\end{itemize}

\begin{table*}[t]
    \begin{center}
    \scalebox{0.91}{
    \setlength\tabcolsep{3.5pt}
    \begin{tabular}{cllccc}
    \toprule[\heavyrulewidth]
    & \multicolumn{1}{c}{\multirow{2}{*}{ Caption (\underline{target}) }} &  \multicolumn{1}{c}{\multirow{2}{*}{ Visual Tags }} & \multirow{2}{*}{ GT } & \multicolumn{2}{c}{Prediction} \\
    \cmidrule(r){5-6}
    & &  &  & (W+C+V) & (W+C)  \\
    \midrule
    \midrule
     \multirow{7}{*}{ + } & \textit{``The \underline{curry}'s  \includegraphics[height=1em]{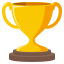}"} & parade, marching, urban area, ... & B-PER & B-PER & O \\
    & \textit{``Grandma w dat lit \underline{Apple} Crisp"} & funnel cake, melting, frozen, ... & O & O & B-ORG \\
     & \textit{``Okay \underline{duke dumont} \includegraphics[height=1em]{figures/emoji/1f60d.png}"} & DJ, guitarist, circus, ... & B,I-PER & B,I-PER & O,O \\                         
     & \textit{``\underline{CSI} with my hubby"} & TV, movie, television, ... & B-MISC & B-MISC & B-ORG \\   
     & \textit{``Twin day at \underline{angel stadium}"} & stadium, arena, stampede, ... & B,I-LOC & B,I-LOC & O,O \\
     & \textit{``LETS GO \underline{CID}"} & drum, DJ, drummer, ... & B-PER & B-PER & O \\         
     & \textit{``\underline{MARSHMELLOOOOOOOOS}"} & DJ, night, martini, ... & B-PER & B-PER & O \\
    \midrule
    \multirow{3}{*}{ - } & \textit{``Y'all come see me at \underline{bojangles}. \includegraphics[height=1em]{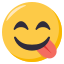}"}
    & floor, tile, airport terminal, ... & B-ORG & O & B-ORG   \\
                         & \textit{``If u're not watching this season of}  & \multirow{2}{*}{monitor, suite, cubicle, ...} & \multirow{2}{*}{B-MISC} & \multirow{2}{*}{O} & \multirow{2}{*}{B-MISC} \\
                         & \textit{\underline{bachelorette} ur doing LIFE WRONG"}  & &  &  &  \\                         
    \bottomrule[\heavyrulewidth]
    \end{tabular}
    }
\end{center}
    \vspace{-5pt}    
   \caption{Error analysis: \textbf{when do images help NER}? Ground-truth labels (GT) and predictions of our model with vision input (W+C+V) and the one without (W+C) for the \underline{underlined} named entities (or false positives) are shown. For interpretability, visual tags (label output of InceptionNet) are presented instead of actual feature vectors used. } \vspace{-11pt}
    \label{tab:wc_vs_wcv}
\end{table*}

\subsection{Results: SnapCaptions Dataset}
\label{subsec:empirical_evaluation:snap_results}
Table \ref{tab:ner_snap_captions} shows the NER performance on the \textit{Snap Captions} dataset.
We report both entity types recognition (PER, LOC, ORG, MISC) and named entity segmentation (named entity or not) results.

\textbf{Parameters}: We tune the parameters of each model with the following search space (bold indicate the choice for our final model): character embeddings dimension: \{25, 50, 100, \textbf{150}, 200, 300\}, word embeddings size: \{25, 50, 100, \textbf{150}, 200, 300\}, LSTM hidden states: \{25, 50, \textbf{100}, 150, 200, 300\}, and $\overline{x}$ dimension: \{25, 50, 100, \textbf{150}, 200, 300\}.
We optimize the parameters with Adagrad \cite{Adagrad} with batch size 10, learning rate 0.02, epsilon $10^{-8}$, and decay 0.0. 

\textbf{Main Results}: When visual context is available (W+C+V), we see that the model performance greatly improves over the textual models (W+C), showing that visual contexts are complimentary to textual information in named entity recognition tasks.
In addition, it can be seen that the modality attention module further improves the entity type recognition performance for (W+C+V).
This result indicates that the modality attention is able to focus on the most effective modality (visual, words, or characters) adaptive to each sample to maximize information gain.
Note that our text-only model (W+C) with the modality attention module also significantly outperform the state-of-the-art baselines \cite{Aguilar+17,Ma+16,Lample+16} that use the same textual modalities (W+C), showing the effectiveness of the modality attention module for textual models as well.

\textbf{Error Analysis}: Table \ref{tab:wc_vs_wcv} shows example cases where incorporation of visual contexts affects prediction of named entities.
For example, the token `curry' in the caption \textit{``The \underline{curry}'s  \includegraphics[height=1em]{figures/emoji/1f3c6.png}"} is polysemous and may refer to either a type of food or a famous basketball player `Stephen Curry', and the surrounding textual contexts do not provide enough information to disambiguate it.
On the other hand, visual contexts (visual tags: `parade', `urban area', ...) provide similarities to the token's distributional semantics from other training examples (\eg snaps from ``NBA Championship Parade Story"), and thus the model successfully predicts the token as a named entity.
Similarly, while the text-only model erroneously predicts `Apple' in the caption \textit{``Grandma w dat lit \underline{Apple} Crisp"} as an organization (\eg Apple Inc.), the visual contexts (describing objects related to food) help disambiguate the token, making the model predict it correctly as a non-named entity (a fruit).
Trending entities (musicians or DJs such as `CID', `Duke Dumont', `Marshmello', etc.) are also recognized correctly with strengthened contexts from visual information (describing concert scenes) despite lack of surrounding textual contexts.
A few cases where visual contexts harmed the performance mostly include visual tags that are unrelated to a token or its surrounding textual contexts.

\begin{figure*}[t]
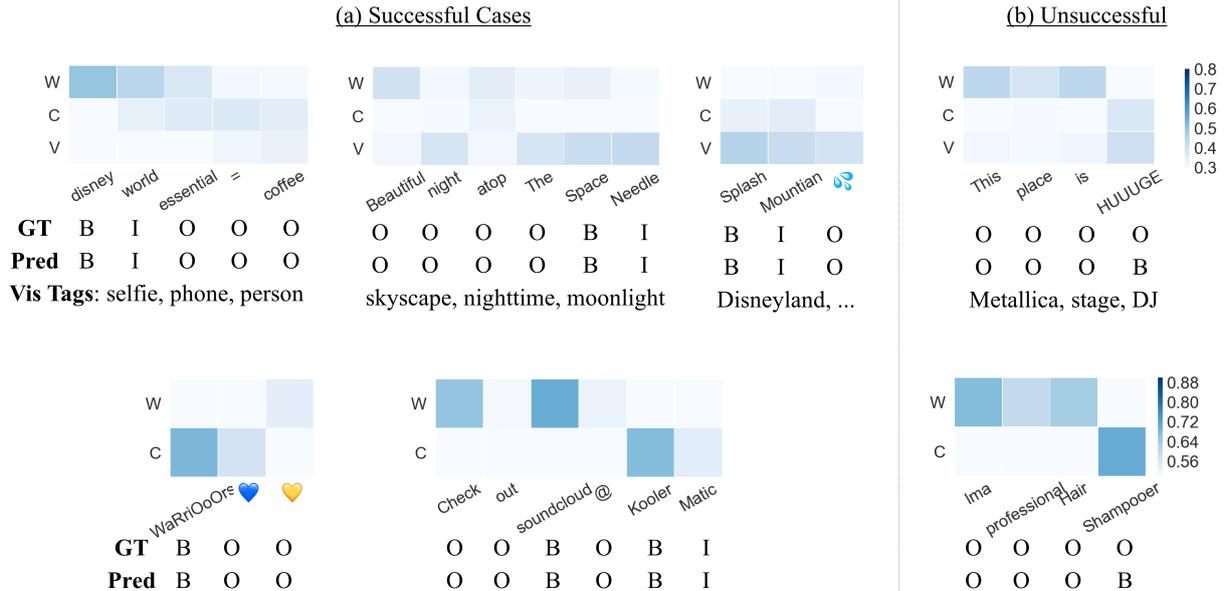

 \include{tex_modules/fig_vis_attention}
  \vspace{-17pt} 
 \caption{ \textbf{Visualization of modality attention} (a) successful cases and (b) unsuccessful ones from \textit{SnapCaptions} test data. For each decoding step of a token (column), the modality attention module amplifies the most relevant modality (darker) while attenuating irrelevant modalities (lighter). The model makes final predictions based on weighted signals from all modalities. For interpretability, visual tags (label output of InceptionNet) are presented instead of actual feature vectors used. GT: ground-truth, Pred: prediction by our model. Modalities- W: words, C: characters, V: visual.  }
  \label{fig:vis_attention}
  \vspace{-6pt}
\end{figure*}

\textbf{Visualization of Modality Attention}: 
Figure \ref{fig:vis_attention} visualizes the modality attention module at each decoding step (each column), where amplified modality is represented with darker color, and attenuated modality is represented with lighter color.

For the image-aided model (W+C+V; upper row in Figure \ref{fig:vis_attention}), we confirm that the modality attention successfully attenuates irrelevant signals (\eg selfies, etc.) and amplifies relevant modality-based contexts in prediction of a given token.
In the example of ``\textit{\underline{disney word} essential = coffee}" with visual tags \textit{selfie, phone, person}, the modality attention successfully attenuates distracting visual signals and focuses on textual modalities, consequently making correct predictions.
The named entities in the examples of ``\textit{Beautiful night atop The \underline{Space Needle}}" and ``\textit{\underline{Splash Mountain}}" are challenging to predict because they are composed of common nouns (space, needle, splash, mountain), and thus they often need additional contexts to correctly predict.
In the training data, visual contexts make stronger indicators for these named entities (space needle, splash mountain), and the modality attention module successfully attends more to stronger signals.

For text-only model (W+C), we observe that performance gains mostly come from the modality attention module better handling tokens unseen during training or unknown tokens from the pre-trained word embeddings matrix.
For example, while \textit{\underline{WaRriOoOrs}} and \textit{\underline{Kooler Matic}} are missing tokens in the word embeddings matrix, it successfully amplifies character-based contexts (\eg capitalized first letters, similarity to known entities `Golden State Warriors') and suppresses word-based contexts (word embeddings for unknown tokens \eg `WaRriOoOrs'), leading to correct predictions.
This result is significant because it shows performance of the model, with an almost identical architecture, can still improve without having to scale the word embeddings matrix indefinitely. 

Figure \ref{fig:vis_attention} (b) shows the cases where the modality attention led to incorrect predictions.
For example, the model predicts missing tokens \textit{\underline{HUUUGE}} and \textit{\underline{Shampooer}} incorrectly as named entities by amplifying misleading character-based contexts (\eg capitalized first letters) or visual contexts (\eg concert scenes, associated contexts of which often include named entities in the training dataset).

\begin{table}[t]
    \begin{center}
    \scalebox{1.00}{
    \setlength\tabcolsep{3.5pt}
    \begin{tabular}{ccc}
    \toprule[\heavyrulewidth]
     Vocab Size & w/o M.A. & w/ M.A. \\
    \midrule
    \midrule
    100\% & 48.8 & \textbf{50.5}  \\
    \midrule
    75\%  & 48.7 & \textbf{50.1}  \\
    50\%  & 47.8 &  \textbf{49.6} \\
    25\%  & 46.4 &  \textbf{48.7} \\
    \bottomrule[\heavyrulewidth]
    \end{tabular}
    }
\end{center}

    \vspace{-10pt}
   \caption{NER performance (F1) on SnapCaptions with \textbf{varying word embeddings vocabulary size}. Models being compared: (W+C) Bi-LSTM/CRF + Bi-CharLSTM w/ and w/o modality attention (M.A.) }    
    \vspace{-16pt}   
    \label{tab:vocab_size}
\end{table}

\textbf{Sensitivity to Word Embeddings Vocabulary Size}:
In order to isolate the effectiveness of the modality attention module on textual models in handling missing tokens, we report the performance with varying word embeddings vocabulary sizes in Table \ref{tab:vocab_size}.
By increasing the number of missing tokens artificially by randomly removing words from the word embeddings matrix (original vocab size: 400K), we observe that while the overall performance degrades, the modality attention module is able to suppress the peformance degradation.
Note also that the performance gap generally gets bigger as we decrease the vocabulary size of the word embeddings matrix.
This result is significant in that the modality attention is able to improve the model more robust to missing tokens without having to train an indefinitely large word embeddings matrix for arbitrarily noisy social media text datasets.

\section{Conclusions}
\label{sec:conclusions}

We proposed a new multimodal NER (MNER: image + text) task on short social media posts.
We demonstrated for the first time an effective MNER system, where visual information is combined with textual information to outperform traditional text-based NER baselines.
Our work can be applied to myriads of social media posts or other articles across multiple platforms which often include both text and accompanying images.
In addition, we proposed the \textit{modality attention} module, a new neural mechanism which learns optimal integration of different modes of correlated information. 
In essence, the modality attention learns to attenuate irrelevant or uninformative modal information while amplifying the primary modality to extract better overall representations.
We showed that the modality attention based model outperforms other state-of-the-art baselines when text was the only modality available, by better combining word and character level information.


\bibliographystyle{acl_natbib}
\bibliography{bibliography}

\end{document}